\documentclass[11pt,letterpaper]{article}

\usepackage[textwidth=6in,textheight=8.5in,centering]{geometry}
\usepackage{booktabs}
\usepackage{graphicx}
\usepackage{amsmath,amssymb}
\usepackage{multirow}
\usepackage{subcaption}
\usepackage{natbib}
\usepackage{url}
\usepackage{xcolor}
\usepackage{hyperref}
\usepackage[load-configurations=version-1]{siunitx}

\title{Flow-of-Thought: A Framework for Visual Reasoning}
\author{Mariia Baidachna \qquad Nicolas Pugeault\\[0.5em]
{\normalsize University of Glasgow}}
\date{}

\begin{document}

\maketitle

\begin{abstract}
Mental imagery, ``seeing with the mind's eye'' is an essential aspect of human cognition. Despite rapid progress Large Language Models (LLMs) and Vision Transformers (ViTs) still underperform on tasks requiring spatial understanding. To address this, we introduce Flow-of-Thought (FoT), a framework that integrates the generation of visual sketches as intermediate reasoning steps, mimicking mental imagery in humans. We train coordinate-aware trajectory flow fields on $SO(2)$ group orbits and cumulative shortest paths, then freeze the learned dynamics; same vs. different decisions compare competing generative hypotheses using foreground-weighted reconstruction energy. On locked tests FoT reaches 100.0\% accuracy on Tetris and 99.0\% on colored shapes. Under frozen transfer, the orbit-trained 2D flow improves over its endpoint-only control on BLINK Multi-view (72.2\% vs. 63.9\% on 133 public validation pairs), supporting continuous visual traces as an effective and interpretable representation for spatial reasoning in some out-of-distribution settings.
\end{abstract}

\section{Introduction}

Modern foundation models have achieved impressive performance at a wide range of tasks, yet they still fail to generalize to spatial planning or visual logic tasks that are often trivial for humans \citep{vsp2025, topobench2026}. One example is an IQ-style question that presents a source block and a target block, which has been rotated or also mirrored, and asks to determine if the blocks are the same. One characteristic of such tasks is that they are difficult to solve in one-shot, but human subjects perform well, mentally simulating incremental rotation steps that lead to the solution \citep{shepard1971mental}. Similarly, it is difficult to determine the path from the start to the end of a maze, but intermediate sketches of the path make the task trivial \citep{lotfi2024cos}. In humans, mental imagery, ``seeing with the mind's eye'' is believed to be essential to solving this type of visual reasoning tasks. Text-based models on the other hand, lack a ``mental scratchpad'' and instead attempt to hold entire spatial states in fixed attention mechanisms, which fails for complex geometries \cite{zhang2024mindseye, latentsketchpad2025}.

Recent advances in intermediate reasoning have significantly contributed to performance gains in text-based tasks \citep{openai2024o1, deepseek2025deepseekr1, lightman2023lets}. Despite the success of large language models (LLMs) with built-in reasoning, these frameworks fail to generalize outside of semantic and symbolic domains \citep{dziri2023faith, fu2024blink, vsp2025}. This limitation stems from a fundamental modality mismatch: text is inherently one-dimensional and discrete, whereas visual reasoning requires understanding multi-dimensional topology, continuous geometric transformations, and spatial relationships. Recent evaluations of state-of-the-art (SoTA) Multimodal Large Language Models (MLLMs) reveal that even when equipped with powerful vision encoders, these systems suffer from a perception-reasoning gap. Tasks that are deeply ingrained in human cognitive core knowledge \citep{spelke2007core}, such as mental rotation or shortest connected path through a maze, often result in severe spatial hallucinations or logical collapse in MLLMs \citep{fu2024blink, vsp2025}. Without the ability to explicitly track geometric changes step-by-step, MLLMs are forced to rely on semantic memorization rather than true spatial thinking \citep{chollet2019arc}, motivating frameworks that can reason through continuous visual states.

Following a neuroscience-inspired view of mental imagery as a continuously evolving internal representation \citep{shepard1971mental,kosslyn1978visual}, we introduce a visual reasoning framework, \textit{Flow-of-Thought (FoT)}, that learns continuous visual dynamics from trajectory supervision using flow matching \citep{lipman2022flow}. A coordinate-, time-, and action-conditioned spatial U-Net predicts dense transport fields along rendered rotation orbits and additive dynamics over cumulative shortest-path masks for mazes. Integrating these fields as ordinary differential equations (ODEs) produces an intermediate visual state at any time $t\in[0,1]$. After model selection, the learned dynamics are frozen. For same-versus-different rotation decisions, the source and its horizontal reflection define competing hypotheses, and the hypothesis with the lower foreground-weighted target-reconstruction energy over a predeclared action set is selected. Thus, the endpoint evidence used for the decision is generated by the same learned trajectories whose intermediate states can be inspected.

We evaluate FoT on locked synthetic rotation tests, maze navigation, and frozen transfer to 3D Blocks and BLINK Multi-view, with comparisons to matched controls, trained vision models, and multimodal language models. FoT reaches 100.0\% accuracy on Tetris and 99.0\% on colored shapes, while its maze traces reach 97.5\% endpoint IoU and 100\% goal reach, although later path segments sometimes activate prematurely. On BLINK Multi-view, the larger of the two transfer evaluations (133 public validation pairs), the orbit-trained flow reaches 72.2\%, compared with 63.9\% for its endpoint-only control; we treat this as the primary transfer result, while noting that it remains a public-validation evaluation whose sign convention was finalized after a two-item smoke check. On 3D Blocks it reaches 61.5\% and does not exceed the endpoint-only control at 66.7\%; because the test set contains only 78 pairs, a single reclassified item moves accuracy by more than 1.2 points and the estimate is correspondingly seed-sensitive, so we report the 3D evaluation as an instability-limited negative control against a universal out-of-distribution (OOD) advantage from intermediate trajectory supervision.

\section{Related Work}

\textbf{Chain-of-Thought (CoT) and iterative reasoning.} Modern reasoning models generate intermediate computations before an answer \citep{deepseek2025deepseekr1, openai2024o1}. AlphaGeometry \citep{trinh2024solving} extended structured reasoning to geometric proofs. More recently, Diffusion of Thought (DoT) \citep{ye2024diffusion} challenged the sequential token paradigm by integrating reasoning into diffusion models, highlighting the benefit of refining intermediate steps globally over time.

\textbf{Visual Reasoning and Chain-of-Sketch.} To address spatial limitations, ``Visual Sketchpad'' \citep{hu2024visual} allowed models to generate code to plot geometric auxiliary lines. Extending this, \cite{lotfi2024cos} formalized ``Inductive CoS,'' generating visual traces frame-by-frame to solve high ``globality degree'' tasks (e.g., mazes). This physical grounding is critical; recent evaluations in the ``Illusion of Thinking'' \cite{shojaee2025illusion} highlight that models often memorize puzzle heuristics rather than adapting dynamically. Without an externalized visual buffer, foundation models suffer from acute epistemic boundaries in spatial tasks.

\textbf{Cognitive Neuroscience of Spatial Logic.} The architecture of FoT is heavily motivated by human psychology. For example, \citet{shepard1971mental} used classic mental rotation experiments to show that human reaction times for verifying chiral objects are proportional to their angular difference, providing evidence that the brain simulates a continuous rotation in the ``mind's eye''. Similarly, \cite{kosslyn1978visual} showed that spatial navigation relies on continuous mental scanning. By modeling our visual transition operator as a continuous ODE, FoT mimics this psychological process.

\section{Preliminaries}

The overarching goal is to create a framework that reasons through interpretable visuals to solve spatial tasks. Formally, we consider visual reasoning datasets $\mathcal{D}$ whose samples contain an input condition $c$, an initial visual state $x_0$, and either a target trajectory $\gamma(t)$ or a final label $o$. A time-conditioned spatial field defines the continuous dynamics
\begin{equation}
    \frac{d x(t)}{d t} = v_\theta(x(t),t,c,a), \qquad t\in[0,1],
\end{equation}
where $a$ is a task action, such as a requested rotation. Unlike linear pixel interpolation, our rotation supervision follows the rendered $SO(2)$ orbit, while maze supervision follows cumulative prefixes of the shortest path. The learned field therefore exposes an arbitrary-time visual state rather than only a final prediction.

\begin{figure*}[p]
  \centering
  \includegraphics[width=.96\textwidth]{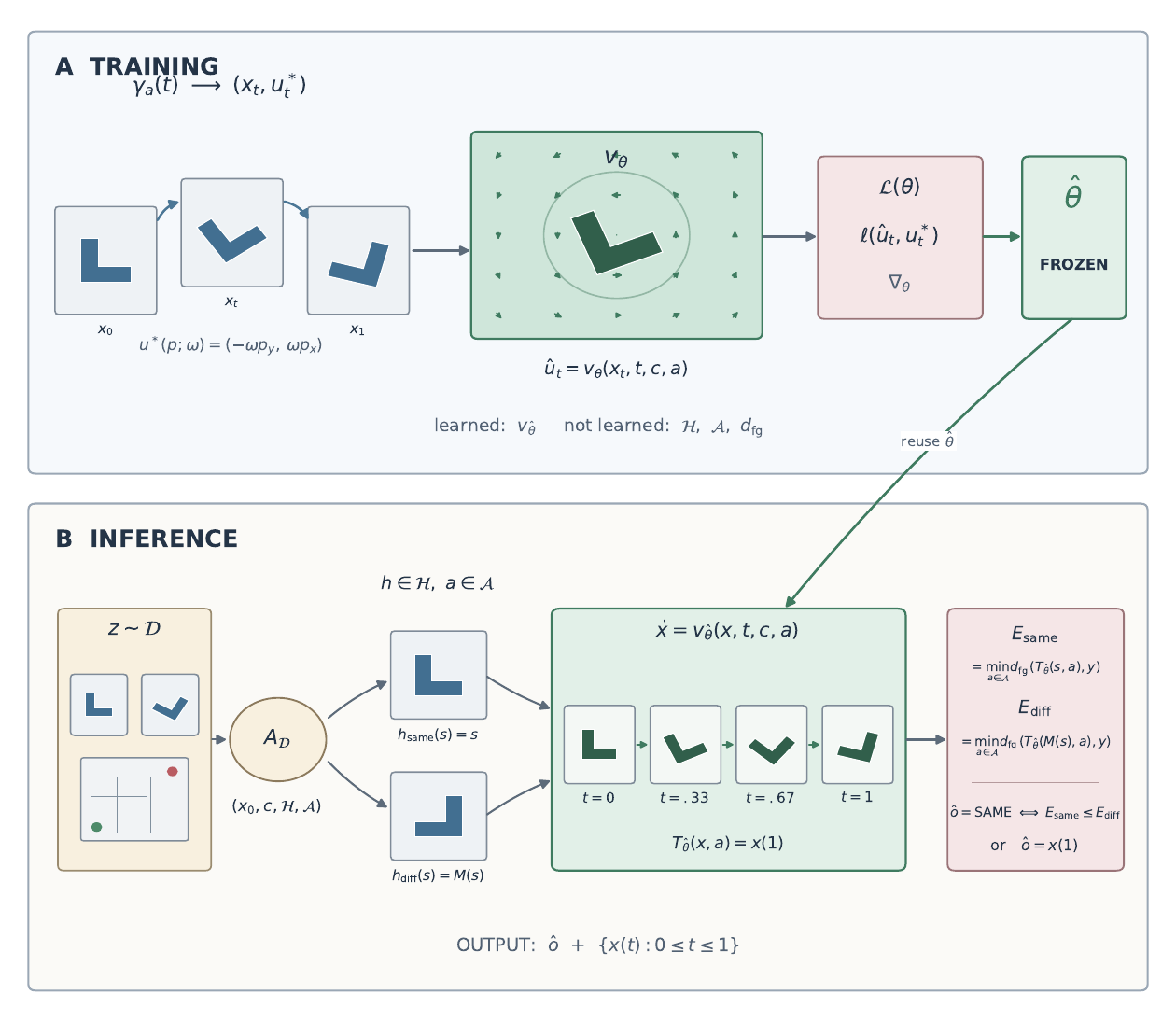}
  \caption{Training and inference in FoT. \textbf{Top:} the rotation example follows the rendered path $x_t=\gamma_a(t)$ with target transport $u^*(p;\omega)=(-\omega p_y,\omega p_x)$. The field predicts $\hat u_t=v_\theta(x_t,t,c,a)$ and minimizes $\ell(\hat u_t,u_t^*)$; thus only $\theta$ is learned. For mazes, the same slots contain the trace $\tau(t)$, map condition $m$, and additive field $v_\theta(\tau(t),t,m)$. \textbf{Bottom:} $A_{\mathcal{D}}$ maps an arbitrary compatible item to $(x_0,c,\mathcal{H},\mathcal{A})$. Each hypothesis/action is integrated under the same frozen field, producing the trace $\{x(t)\}$ and endpoint $T_{\hat\theta}(x,a)=x(1)$. Rotation pairs use the paper's exact SAME/DIFFERENT energies $E_{\rm same}$ and $E_{\rm diff}$; dense tasks return $x(1)$. Only $A_{\mathcal{D}}$ and the readout are dataset-specific.}
  \label{fig:pipeline}
\end{figure*}

Figure~\ref{fig:pipeline} separates learning from frozen inference. For a compatible dataset $\mathcal{D}$, let $A_{\mathcal{D}}(z)=(x_0,c,\mathcal{H},\mathcal{A})$ denote the deterministic adapter from an item $z$ to the model interface: initial state, condition, candidate hypotheses, and allowed actions. For rotation discrimination, $\mathcal{H}=\{h_{\rm same},h_{\rm diff}\}$ with $h_{\rm same}(s)=s$ and $h_{\rm diff}(s)=M(s)$.

\section{The Flow-of-Thought Framework}

\subsection{Trajectory Flow Field}

The shared backbone is a coordinate-aware, time- and action-conditioned spatial U-Net. The current state, the initial/task condition, and explicit $(x,y)$ coordinates enter the convolutional stem. Sinusoidal time features and a three-dimensional rotation action representation $(\Delta\theta/180,\sin\Delta\theta,\cos\Delta\theta)$ are embedded by MLPs and injected by feature-wise linear modulation throughout the residual blocks. Group normalization and SiLU activations are used at every resolution.

For rotation, the network emits a dense two-channel transport field rather than an image residual. Let normalized image coordinates be $(p_x,p_y)$ and $\omega$ the requested angular displacement in radians. At a state rendered on the rotation orbit, the target tangent field is
\begin{equation}
    u^*(p;\omega)=(-\omega p_y,\omega p_x),
\end{equation}
and the local objective is a smooth $\ell_1$ loss between the predicted and target fields. We integrate the learned field with 12 fixed Heun steps. A parallel backward characteristic map is integrated continuously, and every requested image $x(t)$ is sampled once from the original source. This single-source rendering avoids the blur that would accumulate by repeatedly resampling intermediate images.

For mazes, the neural state $\tau(t)$ is a cumulative path mask and the field is additive:
\begin{equation}
    \frac{d\tau(t)}{dt}=v_\theta(\tau(t),t,m),
\end{equation}
where $m$ encodes walls, start, and goal. Training uses foreground-balanced tangent regression together with differentiable multi-step binary-cross-entropy and Dice losses against cumulative shortest-path prefixes. Eight Heun steps produce the final path and its intermediate trace.

\subsection{Frozen Hypothesis Competition}

For a rotation pair $(s,y)$, SAME starts from the observed source $s$ and DIFFERENT starts from its horizontal reflection $M(s)$. Let $T_\theta(x,a)$ be the endpoint produced by the frozen learned transport and let
\begin{equation}
\begin{aligned}
    d_{\mathrm{fg}}(\hat y,y)
    &= \frac{1}{CHW}\sum_{c,p}
       \left(1+4\max_c y_{c,p}\right) \\
    &\qquad\cdot(\hat y_{c,p}-y_{c,p})^2.
\end{aligned}
\end{equation}
The branch energies are nuisance minima over the predeclared 36-angle set
\begin{equation*}
    \mathcal{A}
    = \{-177.5^\circ,-167.5^\circ,\ldots,172.5^\circ\}.
\end{equation*}
For the two competing hypotheses,
\begin{equation}
\begin{aligned}
    E_{\mathrm{same}}
    &= \min_{a\in\mathcal{A}}
       d_{\mathrm{fg}}\!\left(T_\theta(s,a),y\right), \\
    E_{\mathrm{diff}}
    &= \min_{a\in\mathcal{A}}
       d_{\mathrm{fg}}\!\left(T_\theta(M(s),a),y\right).
\end{aligned}
\end{equation}
We predict SAME iff $E_{\mathrm{same}}\leq E_{\mathrm{diff}}$. The $2.5^\circ$ offset keeps candidates off both manifest angle grids. No pair labels, fitted threshold, target-task adaptation, or analytic image transformation enters this decision. External evaluations instantiate the benchmark's declared answer hypotheses: 3D Blocks compares original and reflected sources under the signed pair of the provided angular disparity, while BLINK compares positive and negative apparent object-motion candidates (10--170$^\circ$) corresponding to the two camera-orbit directions.

\section{Data and Experiments}

\begin{figure}[tbp]
  \centering
  \includegraphics[width=.82\linewidth]{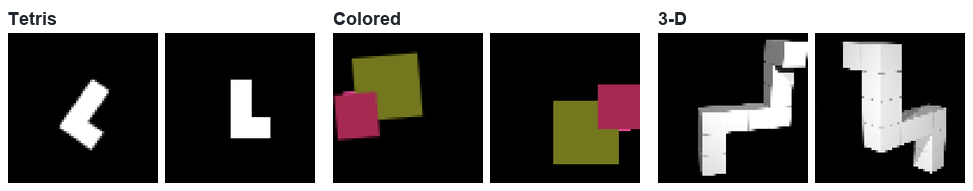}
  \caption{Representative source--target pairs for Tetris, colored shapes, and 3D Blocks \citep{ganis2015new}; each labeled group shows source on the left and target on the right. Maze inputs are shown with their generated traces in Figure~\ref{fig:qualitative_maze}.}
  \label{fig:data_samples}
\end{figure}

We test our framework on the representative rotation tasks in Figure~\ref{fig:data_samples} and on maze navigation. For training details, please refer to Appendix~\ref{app:task_formulations}.

\paragraph{2D Tetris Dataset}
We generate trajectory-supervised ``Tetris-like'' rotations with distinct chiral shapes. The flow is trained on J/L/S/Z sources and selected on held-out F/P shapes, with random initial pose, reflection, and continuous angular displacement. The locked 100-pair test asks whether the target is a rotation of the source or a rotation of its mirror.

\paragraph{Colored Shapes Dataset}
This task uses the same problem formulation as the Tetris dataset but focuses on multi-object compositional rotation and additional RGB channels, increasing the complexity of the spatial state. Training and validation are disjoint deterministic procedural samples, and the reported decision set contains 100 locked pairs.

\paragraph{3D Blocks and BLINK Multi-view}
The 3D Blocks dataset \citep{ganis2015new} contains 78 sample pairs that are rotated or rotated and mirrored, but provides no intermediate states. BLINK Multi-view contributes 133 public validation pairs under a related viewpoint-change decision. We apply the frozen 2D colored-shape flow to both without target-task training. For 3D, the supplied disparity restricts each branch to its signed angle pair; for BLINK, the two answer branches minimize over positive or negative object rotations on a fixed 10$^\circ$ grid up to 170$^\circ$. The BLINK result is exploratory because the evaluation sign convention was finalized after a two-item smoke check. We note the two transfer sets differ in size: at $n{=}78$, a single reclassified 3D pair shifts accuracy by 1.3 points, whereas the BLINK set is roughly 70\% larger.

\paragraph{Maze Navigation Dataset}
Perfect $9\times9$ cell mazes are generated by depth-first backtracking, and breadth-first search supplies the shortest path. The learning target is a sequence of cumulative path masks from the fixed start to the goal, resized to $64\times64$.

\section{Results and Discussion}

\begin{figure}[tbp]
    \centering
    \includegraphics[width=.84\linewidth]{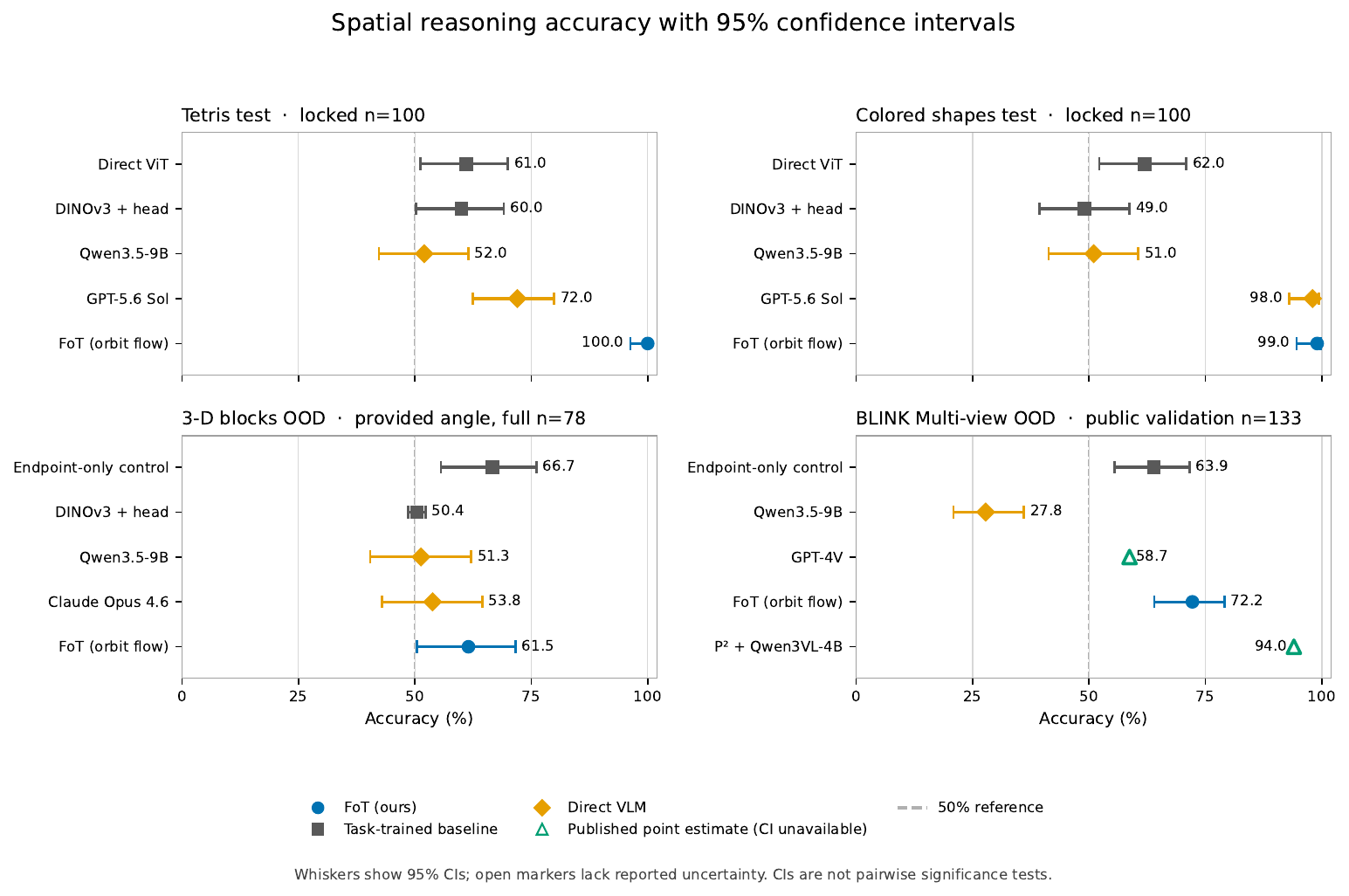}
    \caption{Audited accuracy with 95\% confidence intervals. Open markers are published point estimates without compatible reported uncertainty; intervals are not pairwise significance tests. Maze is omitted because its dense trajectory metrics are not accuracy-equivalent.}
    \label{fig:final_accuracy_ci95}
\end{figure}

Table~\ref{tab:results_all} gives the main comparison across trained vision
models, frontier multimodal models, published systems, and FoT. Dashes denote
unavailable or non-comparable results. The maze rows intentionally preserve the
models' different output interfaces: trace accuracy evaluates a binary
trace-level response, whereas FoT directly generates a path and is evaluated by
endpoint/prefix IoU and path success. Figure~\ref{fig:final_accuracy_ci95}
complements the point-estimate table with uncertainty for the directly
comparable classification results.

\begin{table}[tbp]
  \centering
  \caption{Results (\%). $^\dagger$Published; $^\ddagger$historical items;
  $^\star$incomplete and excluded.}
  \label{tab:results_all}
  \scriptsize
  \setlength{\tabcolsep}{2.6pt}
  \resizebox{\linewidth}{!}{%
  \begin{tabular}{llccccccccc}
    \toprule
    \textbf{Task} & \textbf{Metric} & \textbf{ViT} & \textbf{DINOv3}
    & \textbf{Qwen3.5} & \textbf{GPT-4o} & \textbf{GPT-5.6}
    & \textbf{Claude 4.6} & \textbf{GPT-4V$^\dagger$}
    & \textbf{P$^2$+Qwen3VL$^\dagger$} & \textbf{FoT} \\
    \midrule
    Tetris & Acc. & 61.0 & 60.0 & 52.0 & 52.4$^\ddagger$ & 72.0
      & 46.2$^\ddagger$ & -- & -- & \textbf{100.0} \\
    Colored & Acc. & 62.0 & 49.0 & 51.0 & 50.0$^\ddagger$ & 98.0
      & 51.6$^\ddagger$ & -- & -- & \textbf{99.0} \\
    3D Blocks & Acc. & -- & 50.4 & 51.3 & 50.0 & 57.3$^\star$
      & 53.8 & -- & -- & \textbf{61.5} \\
    BLINK & Acc. & -- & -- & 27.8 & -- & -- & -- & 58.7
      & \textbf{94.0} & 72.2 \\
    \midrule
    \multirow{3}{*}{Maze}
      & Trace acc. & -- & -- & 59.0 & 50.6$^\ddagger$ & 75.0$^\star$
      & 50.0$^\ddagger$ & -- & -- & -- \\
      & End/prefix IoU & -- & -- & -- & -- & -- & -- & -- & --
      & \textbf{97.5/84.2} \\
      & Path success & -- & -- & 0.0 & 0.0$^\ddagger$ & --
      & 0.0$^\ddagger$ & -- & -- & \textbf{100.0} \\
    \bottomrule
  \end{tabular}}
\end{table}

FoT reaches 100.0\% on Tetris and 99.0\% on colored shapes, exceeding the
displayed vision and multimodal baselines. Under frozen transfer, BLINK is the
more informative of the two OOD evaluations: with 133 items,
FoT's 72.2\% [64.0, 79.1] exceeds its endpoint-only control at 63.9\%
[55.5, 71.6] by 8.3 points --- the intervals overlap, so the improvement is
suggestive rather than conclusive --- and exceeds GPT-4V, though not the
published P$^2$+Qwen3VL result. On 3D Blocks, FoT is the strongest complete
entry shown in Table~\ref{tab:results_all}, but this does not establish a
trajectory-supervision advantage: the matched endpoint-only control reported in
the ablations reaches 66.7\%, and the evaluation itself is unstable at this
sample size --- with $n{=}78$, the Wilson interval [50.4, 71.6] spans more than
twenty points, a single reclassified pair moves accuracy by 1.3 points, and the
ranking among nearby methods is seed-sensitive. The evidence therefore supports
trajectory learning on matched 2D tasks and promising transfer on BLINK; the
small, unstable 3D evaluation delimits any universal OOD claim
without precisely measuring the size of the deficit.

For maze navigation, FoT obtains endpoint IoU $97.5$\% [94.6, 100.0],
intermediate prefix IoU $84.2$\% [82.2, 86.3], 100\% goal reach, and zero
obstacle violations. These generation metrics should not be compared directly
with the trace-accuracy row. Moreover, strong endpoint and prefix metrics do
not imply perfectly causal drawing: premature activation is 31.7\% and mean
future-path intensity is 0.164. The model reliably produces a valid path but
sometimes reveals later path segments too early.

\subsection{Qualitative Analysis}

\begin{figure}[tbp]
    \centering
    \includegraphics[width=.92\linewidth]{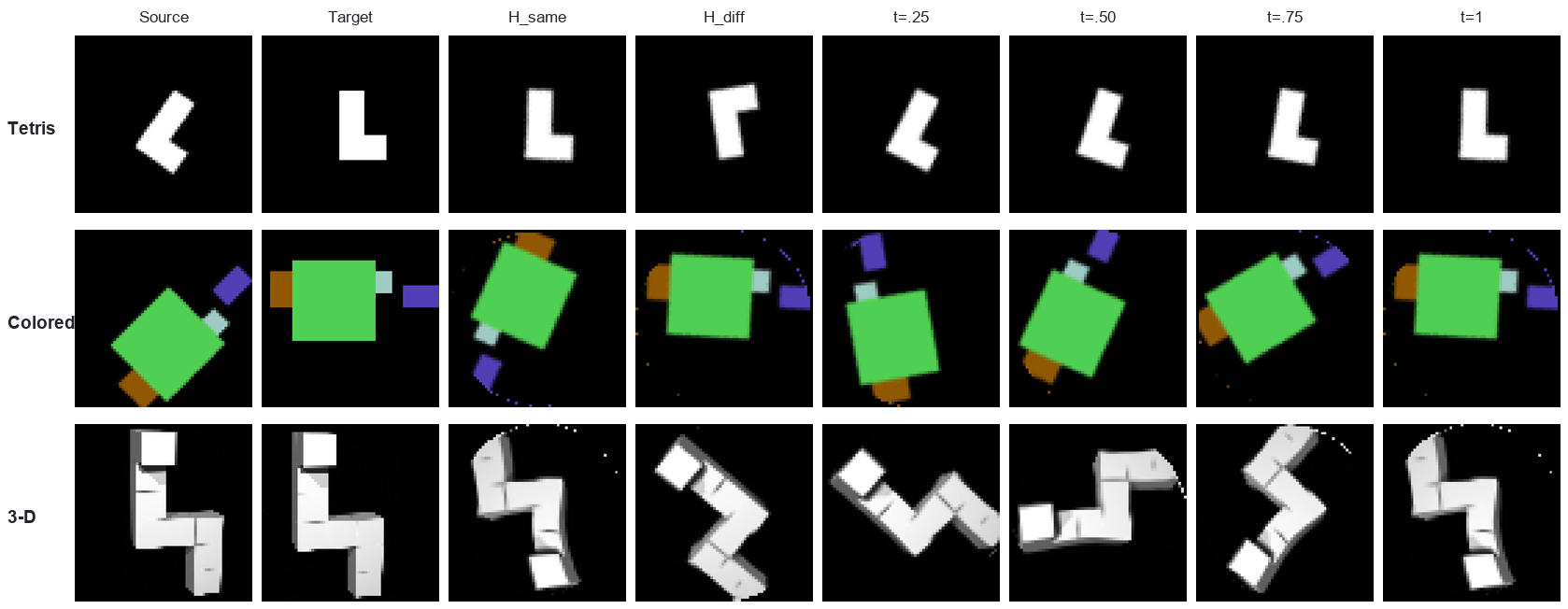}
    \caption{Frozen-flow audits: (a) Tetris SAME, (b) colored-shape DIFFERENT, and (c) 3D Blocks transfer. $H_{\mathrm{same}}$ and $H_{\mathrm{diff}}$ are the best endpoint reconstructions for the competing branches; the remaining columns show the chosen branch at arbitrary times.}
    \label{fig:qualitative_rotation}
\end{figure}

Figure~\ref{fig:qualitative_rotation} combines the rotation-like cases into one space-efficient audit. The SAME and DIFFERENT endpoint hypotheses are shown beside the selected branch trajectory. On the synthetic tasks the intermediate frames preserve object identity while rotating toward the target. On 3D Blocks, the 2D flow often behaves more like viewpoint-conditioned morphing, consistent with the weak and unstable quantitative transfer and the stronger endpoint-only control.

\begin{figure}[tbp]
    \centering
    \includegraphics[width=.90\linewidth]{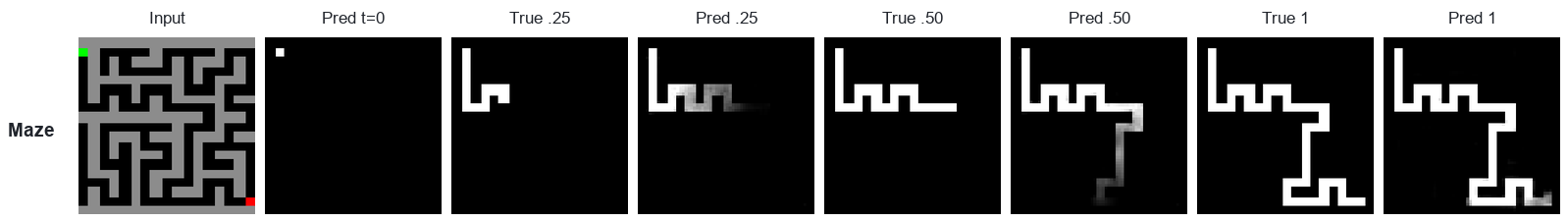}
    \caption{Maze trajectory audit on a held-out maze. Alternating ground-truth and predicted states show accurate cumulative path construction together with faint premature activation.}
    \label{fig:qualitative_maze}
\end{figure}

Figure~\ref{fig:qualitative_maze} places truth and prediction at matched times. The endpoint is nearly exact, while faint early activation of later path segments makes the temporal limitation directly inspectable.

\subsection{Interpretability}
A practical benefit of FoT is that it produces interpretable intermediate images tied directly to its endpoint evidence. In successful rotation decisions, intermediate frames show progressive geometric alignment; in failures, they reveal loss of structure or an implausible branch reconstruction. Maze traces similarly expose whether errors concern wall avoidance, endpoint completion, or the timing of path activation.

\section{Conclusion and Future Work}
This work introduces the FoT framework, bridging the gap between spatial processing and foundation models. A trajectory-supervised transport field provides an interpretable mental scratchpad, while frozen hypothesis competition turns its generated endpoints into decisions. FoT solves the locked synthetic rotation tasks and transfers positively to BLINK, which we regard as the primary OOD evidence, while the small and seed-sensitive 3D evaluation and the maze timing audit mark where the trajectory claim does not yet hold. Future work will target physically faithful 3D dynamics evaluated on a larger 3D test set, stricter causal path generation, and integration of the visual FoT pipeline as an external tool for multimodal foundation models. The code will be made publicly available upon acceptance.

\bibliographystyle{plainnat}
\bibliography{main}

@article{lotfi2024cos,
  title={Chain-of-Sketch: Enabling Global Visual Reasoning},
  author={Lotfi, Aryo and Fini, Enrico and Bengio, Samy and Nabi, Moin and Abbe, Emmanuel},
  journal={arXiv preprint arXiv:2410.08165},
  year={2024},
  url={https://arxiv.org/abs/2410.08165}
}

@inproceedings{ye2024diffusion,
  title={Diffusion of Thought: Chain-of-Thought Reasoning in Diffusion Language Models},
  author={Ye, Jiacheng and Gong, Shansan and Chen, Liheng and Zheng, Lin and Gao, Jiahui and Shi, Han and Wu, Chuan and Jiang, Xin and Li, Zhenguo and Bi, Wei and Kong, Lingpeng},
  booktitle={Advances in Neural Information Processing Systems},
  volume={37},
  pages={105345--105374},
  year={2024}
}

@inproceedings{hu2024visual,
  title={Visual Sketchpad: Sketching as a Visual Chain of Thought for Multimodal Language Models},
  author={Hu, Yuzhe and others},
  booktitle={Advances in Neural Information Processing Systems},
  year={2024}
}

@article{lipman2022flow,
  title={Flow Matching for Generative Modeling},
  author={Lipman, Yaron and Chen, Ricky TQ and Ben-Hamu, Heli and Nickel, Maximilian and Le, Matt},
  journal={arXiv preprint arXiv:2210.02747},
  year={2022}
}

@article{latentsketchpad2025,
  title={Latent Sketchpad: Sketching Visual Thoughts to Elicit Multimodal Reasoning in {MLLMs}},
  author={Zhang, Huanyu and Wu, Wenshan and Li, Chengzu and Shang, Ning and Xia, Yan and Huang, Yangyu and Zhang, Yifan and Dong, Li and Zhang, Zhang and Wang, Liang and Tan, Tieniu},
  journal={arXiv preprint arXiv:2510.24514},
  year={2025}
}

@inproceedings{zhang2024mindseye,
  title={Mind's Eye of {LLMs}: Visualization-of-Thought Elicits Spatial Reasoning in Large Language Models},
  author={Wu, Wenshan and Zhang, Yadong and others},
  booktitle={Advances in Neural Information Processing Systems},
  volume={37},
  year={2024}
}

@article{vsp2025,
  title={{VSP}: Assessing the Dual Challenges of Perception and Reasoning in Spatial Planning Tasks for {MLLMs}},
  author={Wu, Qiucheng and Zhao, Handong and Saxon, Michael and Bui, Trung and Wang, William Yang and Zhang, Yang and Chang, Shiyu},
  journal={arXiv preprint arXiv:2407.01863},
  note={Preprint available on OpenReview},
  year={2024},
  url={https://arxiv.org/pdf/2407.01863}
}

@article{topobench2026,
  title={TopoBench: Benchmarking LLMs on Hard Topological Reasoning},
  author={Maniparambil, Mayug and Hoehing, Nils and Kapuriya, Janak and Karuvally, Arjun and Rushe, Ellen and Ventresque, Anthony and O'Connor, Noel and Reid, Fergal},
  journal={arXiv preprint arXiv:2603.12133},
  year={2026}
}

@article{ganis2015new,
  title={A New Set of Three-Dimensional Shapes for Investigating Mental Rotation Processes: Validation Data and Stimulus Set},
  author={Ganis, Giorgio and Kievit, Rogier A},
  journal={Journal of Open Psychology Data},
  volume={3},
  number={1},
  pages={e3},
  year={2015},
  publisher={Ubiquity Press},
  doi={10.5334/jopd.ai}
}

@article{shepard1971mental,
  title={Mental rotation of three-dimensional objects},
  author={Shepard, Roger N and Metzler, Jacqueline},
  journal={Science},
  volume={171},
  number={3972},
  pages={701--703},
  year={1971},
  publisher={American Association for the Advancement of Science}
}

@article{kosslyn1978visual,
  title={Visual images preserve metric spatial information: Evidence from studies of image scanning},
  author={Kosslyn, Stephen M and Ball, Thomas M and Reiser, Brian J},
  journal={Journal of Experimental Psychology: Human Perception and Performance},
  volume={4},
  number={1},
  pages={47},
  year={1978},
  publisher={American Psychological Association}
}

@article{chollet2019arc,
  title={On the Measure of Intelligence},
  author={Chollet, Fran{\c{c}}ois},
  journal={arXiv preprint arXiv:1911.01547},
  year={2019}
}

@article{trinh2024solving,
  title={Solving olympiad geometry without human demonstrations},
  author={Trinh, Trieu H and Wu, Yuhuai and Le, Quoc V and He, He and Luong, Thang},
  journal={Nature},
  volume={625},
  number={7995},
  pages={476--482},
  year={2024},
  publisher={Nature Publishing Group UK London}
}

@article{deepseek2025deepseekr1,
  title={DeepSeek-R1: Incentivizing Reasoning Capability in LLMs via Reinforcement Learning},
  author={DeepSeek-AI},
  journal={arXiv preprint arXiv:2501.12948},
  year={2025}
}

@article{openai2024o1,
  title={OpenAI o1 System Card},
  author={OpenAI},
  journal={OpenAI Technical Report},
  year={2024}
}

@article{shojaee2025illusion,
  title={The Illusion of Thinking: Understanding the Strengths and Limitations of Reasoning Models via the Lens of Problem Complexity},
  author={Shojaee, Paria and others},
  journal={Apple Machine Learning Research},
  year={2025}
}

@article{lightman2023lets,
  title={Let's verify step by step},
  author={Lightman, Hunter and Kosaraju, Vineet and Burda, Yura and Edwards, Harri and Baker, Bowen and Lee, Teddy and Leike, Jan and Schulman, John and Sutskever, Ilya and Cobbe, Karl},
  journal={arXiv preprint arXiv:2305.20050},
  year={2023}
}

@article{fu2024blink,
  title={Blink: Multimodal large language models can see but not perceive},
  author={Fu, Xingyu and Hu, Yushi and Li, Bangzheng and Feng, Yu and Wang, Haoyu and Lin, Xudong and Roth, Dan and Smith, Noah A and Chang, Wei-Chiu and Wang, William Yang},
  journal={arXiv preprint arXiv:2404.10501},
  year={2024}
}

@article{spelke2007core,
  title={Core knowledge},
  author={Spelke, Elizabeth S and Kinzler, Katherine D},
  journal={Developmental science},
  volume={10},
  number={1},
  pages={89--96},
  year={2007},
  publisher={Wiley Online Library}
}

@article{dziri2023faith,
  title={Faith and fate: Limits of transformers on compositionality},
  author={Dziri, Nouha and Lu, Ximing and Sclar, Melanie and Li, Xiang Lorraine and Jian, Liwei and Lin, Bill Yuchen and West, Peter and Bhagavatula, Chandra and Bras, Ronan Le and Hwang, Jena D and others},
  journal={Advances in Neural Information Processing Systems},
  volume={36},
  pages={18046--18072},
  year={2023}
}

\appendix

\section{Technical appendices and supplementary material}
\label{app:technical_appendix}

\subsection{Task-Specific Formulations}
\label{app:task_formulations}

All primary models are trained independently with seeds 0, 1, and 2 on Metal Performance Shaders. Each seed uses 3,000 training trajectories and 400 fixed validation trajectories for 30 epochs, batch size 20, AdamW with weight decay $10^{-4}$, and cosine learning-rate decay. Rotation models use width 16, context dimension 64, 12 integration steps, and learning rate $10^{-3}$; the maze model uses width 32, context dimension 128, 8 steps, and learning rate $2\times10^{-4}$. Checkpoints are selected by validation silhouette IoU for rotation and endpoint path IoU for maze.

\paragraph{Rotation trajectories.}
Every training source is randomly pre-rotated and may be mirrored. A requested continuous displacement defines the full $SO(2)$ orbit, and every supervision frame is rendered from a common base image rather than from the previous frame. The dense tangent target is expressed as a spatial velocity in normalized coordinates. During inference, Heun integration advances both the recurrent field state and a backward sampling map; displayed frames use the latter to sample the original source once at the requested $t$.

\paragraph{Maze trajectories.}
The state is a one-channel cumulative trace, while the three-channel condition encodes walls, start, and goal. The tangent loss weights newly added path pixels by 20 and the current foreground by 3. Every second training batch also receives a differentiable multi-step binary-cross-entropy plus soft-Dice loss, weighted by 2, over all predicted prefixes.

\paragraph{Decision protocol.}
Tetris and colored-shape results use the identical locked 100-item sets used for item-level baselines. The reported score is the normalized margin $(E_{\mathrm{diff}}-E_{\mathrm{same}})/(E_{\mathrm{diff}}+E_{\mathrm{same}})$, and zero is fixed by the generative decision rule. The 3D evaluation uses all 78 provided-angle pairs and compares original/reflected reconstructions at the signed supplied disparity. BLINK uses all 133 public Multi-view validation items and predicts camera direction from the reconstruction margin between positive and negative apparent object rotations. Wilson intervals quantify uncertainty over items, while seed-level geometric metrics use two-sided Student-$t$ intervals over three independent runs.

\subsection{Ablation Studies}
\label{app:ablations}

\begin{figure}[tbp]
    \centering
    \includegraphics[width=.88\linewidth]{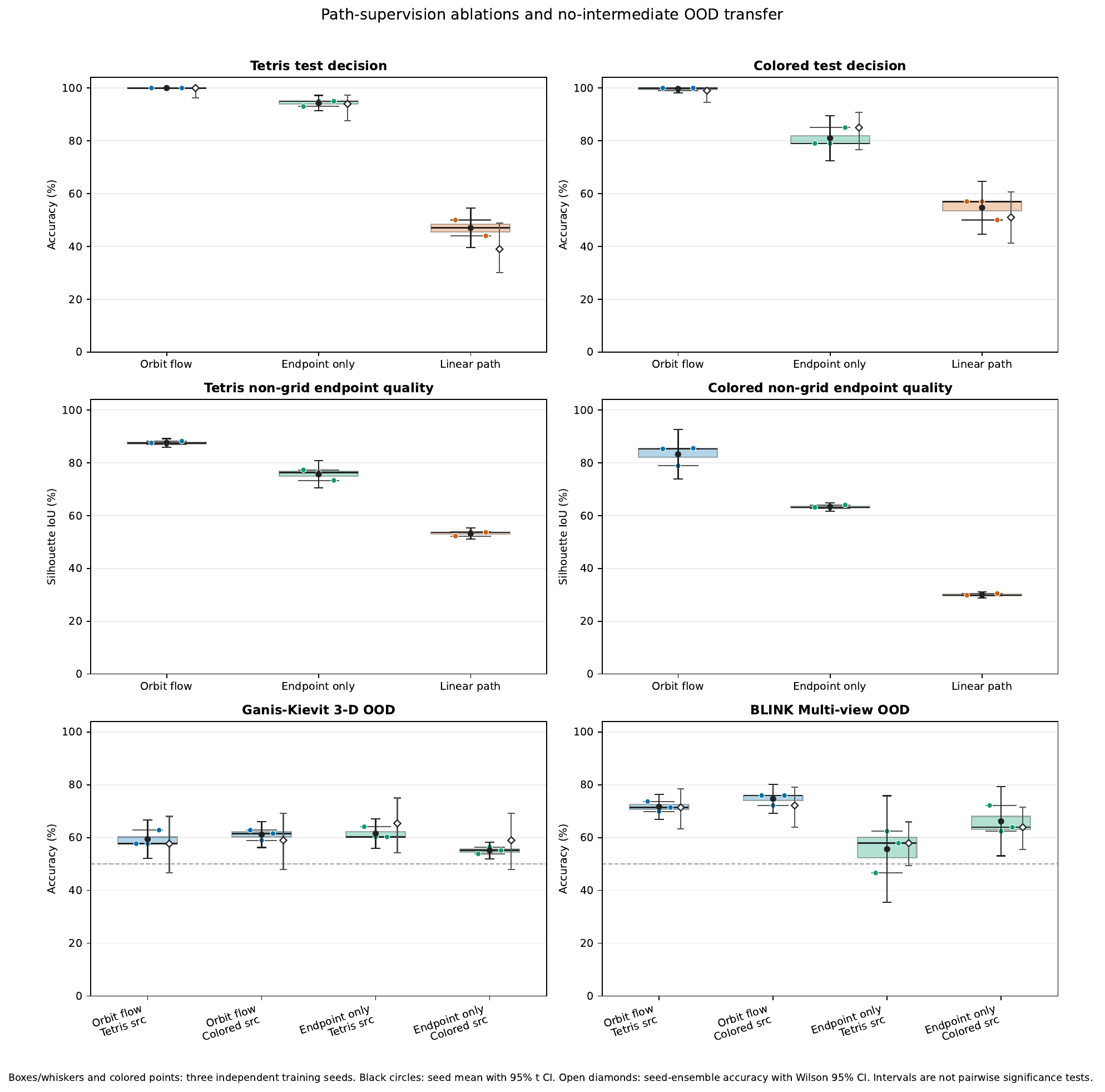}
    \caption{Matched path ablations and no-intermediate OOD controls. Boxes span three training seeds and retain all raw points; black circles and bars denote seed means and two-sided Student-$t$ 95\% intervals. Open diamonds on accuracy panels denote seed-ensemble decisions with Wilson intervals over items.}
    \label{fig:path_ablation}
\end{figure}

The matched supervision ablation holds the source examples, architecture family, optimizer schedule, update count, seeds, frozen reconstruction decision, and action grid fixed. The endpoint-only model sees only final transforms and has no learned state for $0<t<1$; the linear model follows a direct pixel interpolation. Table~\ref{tab:path_ablation} and Figure~\ref{fig:path_ablation} show that rotation-orbit supervision improves both decisions and geometric endpoint quality.

\begin{table}[tbp]
    \centering
    \small
    \setlength{\tabcolsep}{5pt}
    \caption{Matched path-supervision ablation. Accuracy is the seed-ensemble result with a Wilson 95\% interval; endpoint IoU is measured at held-out non-grid angles.}
    \label{tab:path_ablation}
    \begin{tabular}{@{}llcc@{}}
        \toprule
        \textbf{Task} & \textbf{Supervision} & \textbf{Accuracy (\%)} & \textbf{Endpoint IoU} \\
        \midrule
        Tetris & Rotation orbit & \textbf{100.0 [96.3, 100.0]} & \textbf{0.876} \\
        Tetris & Endpoint only & 94.0 [87.5, 97.2] & 0.757 \\
        Tetris & Linear pixel path & 39.0 [30.0, 48.8] & 0.532 \\
        Colored & Rotation orbit & \textbf{99.0 [94.6, 99.8]} & \textbf{0.832} \\
        Colored & Endpoint only & 85.0 [76.7, 90.7] & 0.633 \\
        Colored & Linear pixel path & 51.0 [41.3, 60.6] & 0.300 \\
        \bottomrule
    \end{tabular}
\end{table}

The no-intermediate OOD control gives the required boundary condition. On provided-angle 3D Blocks, orbit flow scores 61.5\% [50.4, 71.6] from colored sources versus 66.7\% [55.6, 76.1] for endpoint-only. On BLINK, orbit flow scores 72.2\% [64.0, 79.1] versus 63.9\% [55.5, 71.6], an 8.3-point improvement. Both transfer sets are modest in size, and the two evaluations should not be weighted equally: on 3D Blocks ($n{=}78$) a single reclassified item moves accuracy by 1.3 points, the two conditions' intervals overlap substantially, and per-seed accuracies vary enough to reorder nearby methods, so the 3D comparison is unstable at this sample size. The BLINK comparison ($n{=}133$) is larger and directionally consistent across seeds, though its intervals also overlap. Intermediate supervision therefore helps on BLINK and shows no advantage on 3D Blocks, with the latter measured too coarsely to quantify the size of the deficit.

\begin{table}[tbp]
    \centering
    \small
    \setlength{\tabcolsep}{4pt}
    \caption{Frozen approximate $SO(2)$ consistency audit. Lower defects are better. The analytic renderer supplies endpoint ground truth only and is not part of the decision.}
    \label{tab:so2_ablation}
    \begin{tabular}{@{}llccc@{}}
        \toprule
        \textbf{Task} & \textbf{Supervision} & \textbf{Identity MSE} & \textbf{Inverse MSE} & \textbf{Composition MSE} \\
        \midrule
        Tetris & Rotation orbit & \textbf{0.000987} & \textbf{0.002526} & \textbf{0.001138} \\
        Tetris & Endpoint only & 0.001061 & 0.017123 & 0.012587 \\
        Tetris & Linear path & 0.011518 & 0.031338 & 0.044970 \\
        Colored & Rotation orbit & 0.001899 & \textbf{0.009159} & \textbf{0.004895} \\
        Colored & Endpoint only & \textbf{0.001855} & 0.030288 & 0.014789 \\
        Colored & Linear path & 0.009657 & 0.053068 & 0.026190 \\
        \bottomrule
    \end{tabular}
\end{table}

\begin{figure}[tbp]
    \centering
    \includegraphics[width=.86\linewidth]{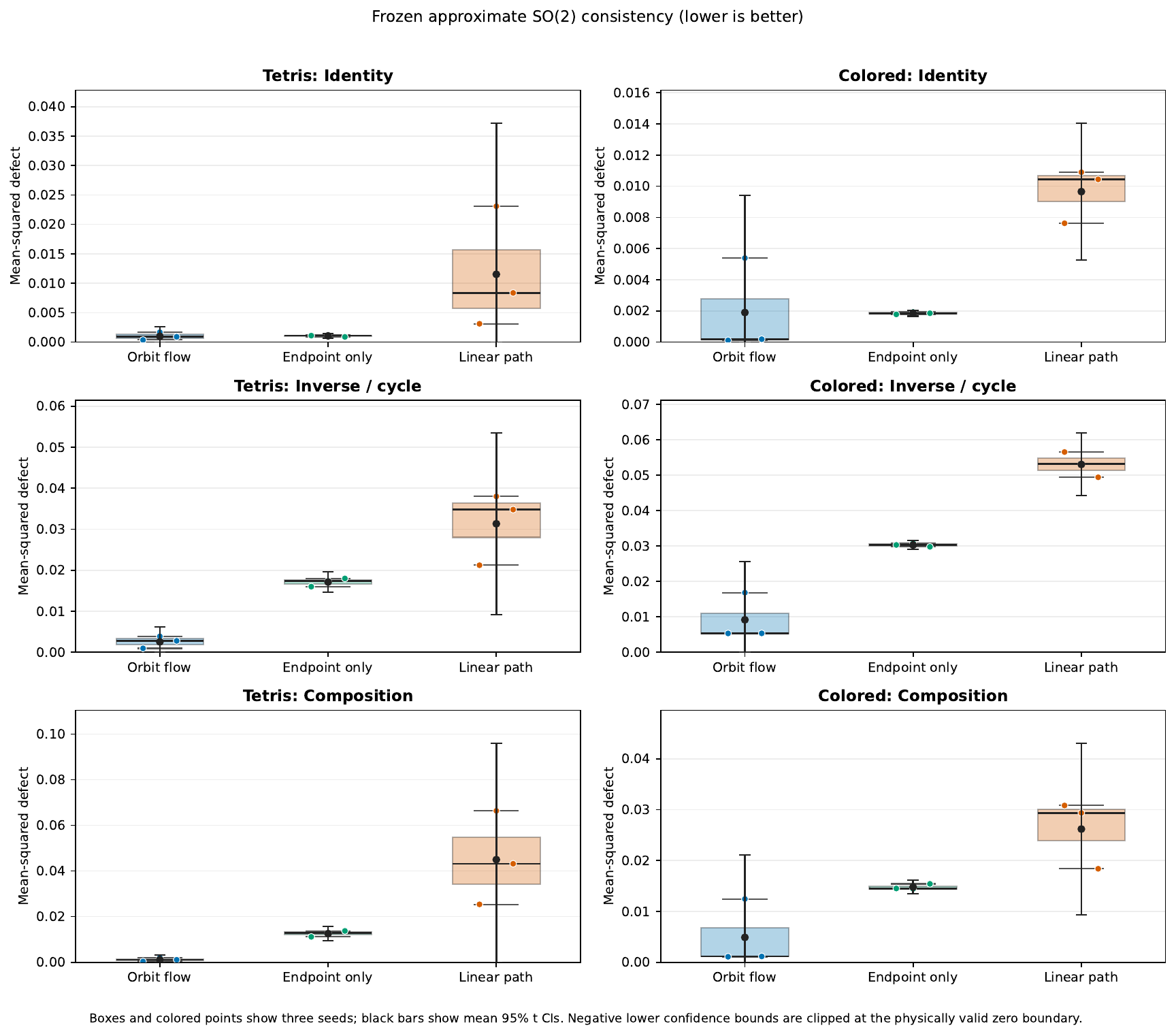}
    \caption{Seed distributions for identity, inverse/cycle, composition, and non-grid endpoint defects. The model is approximately group-consistent; equivariance is not built into the architecture.}
    \label{fig:so2_uncertainty}
\end{figure}

\end{document}